\documentclass[fleqn,10pt]{article}
\usepackage[utf8]{inputenc}
\usepackage[T1]{fontenc}
\usepackage{authblk}      % Para gestionar autores e instituciones

\usepackage{lineno}
\usepackage{tikz}
\usepackage{todonotes}
\usepackage{pgfplots}
\usepackage{array} % Ensure array package is loaded
\usepackage{svg}
\usepackage{graphicx}

\usepackage{caption}
\usepackage{subcaption}

\usepackage{float}
\usepackage{geometry}
\usepackage[acronym]{glossaries}
\setacronymstyle{long-sc-short}
\newacronym{hmm}{HMM}{Human Muscular Manipulability}
\newacronym{kmi}{KMI}{Kinematic Manipulability Index}
\newacronym{dmi}{DMI}{Dynamic Manipulability Index}
\newacronym{lci}{LCI}{Local Conditioning Index}
\newacronym{emg}{EMG}{Electromyography}
\newacronym{mae}{MAE}{Mean Absolute Error}
\newacronym{ae}{AE}{Absolute Error}
\newacronym{imu}{IMU}{Intertial Mesaurement Unit}

\title{CADDI: An in-Class Activity Detection Dataset using IMU data from low-cost sensors}

\author[1]{Luis Marquez-Carpintero}
\author[1]{Sergio Suescun-Ferrandiz}
\author[1]{Monica Pina-Navarro}
\author[1]{Miguel Cazorla}
\author[1]{Francisco Gomez-Donoso\thanks{Corresponding author: Luis Marquez-Carpintero (luis.marquez@ua.es)}}

\affil[1]{Institute for Computer Research, P.O. Box 99. 03080, Alicante, Spain.}

\date{}

\begin{document}
\maketitle

\begin{abstract}

The monitoring and prediction of in-class student activities is of paramount importance for the comprehension of engagement and the enhancement of pedagogical efficacy. The accurate detection of these activities enables educators to modify their lessons in real time, thereby reducing negative emotional states and enhancing the overall learning experience. To this end, the use of non-intrusive devices, such as inertial measurement units (IMUs) embedded in smartwatches, represents a viable solution. The development of reliable predictive systems has been limited by the lack of large, labeled datasets in education. To bridge this gap, we present a novel dataset for in-class activity detection using affordable IMU sensors. The dataset comprises 19 diverse activities, both instantaneous and continuous, performed by 12 participants in typical classroom scenarios. It includes accelerometer, gyroscope, rotation vector data, and synchronized stereo images, offering a comprehensive resource for developing multimodal algorithms using sensor and visual data. This dataset represents a key step toward scalable solutions for activity recognition in educational settings.

\end{abstract}
\newpage

\pgfplotsset{compat=1.18}
\flushbottom
\maketitle

\section*{Background \& Summary}

%(700 words maximum) An overview of the study design, the assay(s) performed, and the created data, including any background information needed to put this study in the context of previous work and the literature. The section should also briefly outline the broader goals that motivated the creation of this dataset and the potential reuse value. We also encourage authors to include a figure that provides a schematic overview of the study and assay(s) design. The Background \& Summary should not include subheadings. This section and the other main body sections of the manuscript should include citations to the literature as needed. 

The ability to estimate the action a human is performing is really useful for a variety of tasks, ranging from healthcare and marketing to monitoring and user interfaces. This task can be carried out using different technologies such as deep learning algorithms and transformer-based architectures. These novel approaches have reportedly very good accuracy but they need vast amounts of labeled data. The data can be sourced from different sensors, for instance color cameras, depth cameras or motion-capture setups. One of the most convenient sensors to gather data to feed human activity prediction systems is the \Gls{imu}. These sensors are able to measure the velocity and orientation. In addition to that, they are present in the majority of smartphones and smartwatches of the industry. Consequently, if an algorithm is developed to predict a user's activity based on the internal IMU of a smartwatch, it has the potential to be widely applicable and beneficial for the majority of smartwatch users. With this in mind, we have created CADDI, an in-Class Activity Detection Dataset using \Gls{imu} data from low-cost sensors. We have captured acceleration and orientation from a regular smartwatch of 12 subjects performing 19 different human actions that can be carried out during a class. The samples were recorded with a closeup stereo color camera, which are also included. The actions considered are continuous (carried out over time), such as resting, drawing or using a keyboard; and instantaneous (happen on a small time frame), such as raising a hand, drinking or turn a page. As stated, the recorded actions are those likely to happen in a classroom.
\\

Several datasets have been developed to predict human actions from \Gls{imu} data. The most relevant ones include watchHAR \cite{ada_alevizaki_2022_7092553}, which contains inconsistent \Gls{imu} data from 20 subjects across 16 categories. VIDIMU \cite{MartnezZarzuela2023VIDIMUMV} is more complete, with \Gls{imu} and image data from 16 subjects in 13 activities, although it uses custom sensors that limit real-world application. PAMAP2 \cite{misc_pamap2_physical_activity_monitoring_231} involves 9 subjects performing 12 activities, with some performing up to 18, but does not specify handedness. Capture-24 \cite{chan2021a} documents a normal day of participants with broadly defined activities, labeled by the subjects themselves, and includes egocentric images. mHEALTH \cite{misc_mhealth_dataset_319} records 12 physical activities from 10 individuals over a short timeframe. OPPORTUNITY AR \cite{misc_opportunity_activity_recognition_226} and OPPORTUNITY++ \cite{yax2-ge53-21} simulate daily activities in a sensor-equipped room with 4 individuals; the newer version includes side-view videos and AI-extracted kinematic data. GOTOV \cite{https://doi.org/10.4121/12716081.v2} focuses on calorimetry in older adults, featuring 35 individuals over 60 years old in 16 activities. REALDISP (REAListic sensor DISPlacement) \cite{misc_realdisp_activity_recognition_dataset_305} investigates sensor displacement effects in activity recognition with 33 categories of physical activity. Finally, HAR \cite{misc_human_activity_recognition_using_smartphones_240} includes 6 continuous physical activities performed by 30 participants using smartphone-embedded IMU.

As it can be seen in this review of related works, the most complete datasets are captured with expensive and complex sensors and only depict physical activities like walking, jumping or jogging. Some other datasets that were captured with regular smartphones or smartwatches are either of physical activities as well or incomplete, providing different categories for each individual. In addition to that, most datasets do not include images of the actions. Finally it is worth noting that, there is no dataset that includes \Gls{imu} data on a classroom environment, and in general, the reviewed datasets focus on activities of daily living and physical exercises.

We focused in the activities carried out in class because it could help to develop algorithms to monitor the activities of students during a lesson. This goal is related to detect stressful, boring or interesting parts so the teachers could be aware in real time and introduce changes in the lessons to reduce negative experiences and improve the engagement in class.

\section*{Methods}
\label{sec:Methods}

%The Methods should include detailed text describing any steps or procedures used in producing the data, including full descriptions of the experimental design, data acquisition assays, and any computational processing (e.g. normalization, image feature extraction). See the detailed section in our submission guidelines for advice on writing a transparent and reproducible methods section. Related methods should be grouped under corresponding subheadings where possible, and methods should be described in enough detail to allow other researchers to interpret and repeat, if required, the full study. Specific data outputs should be explicitly referenced via data citation (see Data Records and Citing Data, below).

%Authors should cite previous descriptions of the methods under use, but ideally the method descriptions should be complete enough for others to understand and reproduce the methods and processing steps without referring to associated publications. There is no limit to the length of the Methods section. Subheadings should not be numbered.

This section describes the data capture process and the custom setup used to collect data for in-class activity detection. The rig includes a stereo camera, smartwatch, and an a computer. Data from the smartwatch sensors, including accelerometer, gyroscope, and rotation vector, are collected and synchronized with stereo camera images. The protocol differentiates between continuous and instantaneous activities, ensuring accurate and synchronized data collection for a wide range of classroom actions.

\subsection*{Capture Setup}
\label{subsec:capture-setup}

\begin{figure}[H]
    \centering
    \includegraphics[width=1.0\textwidth]{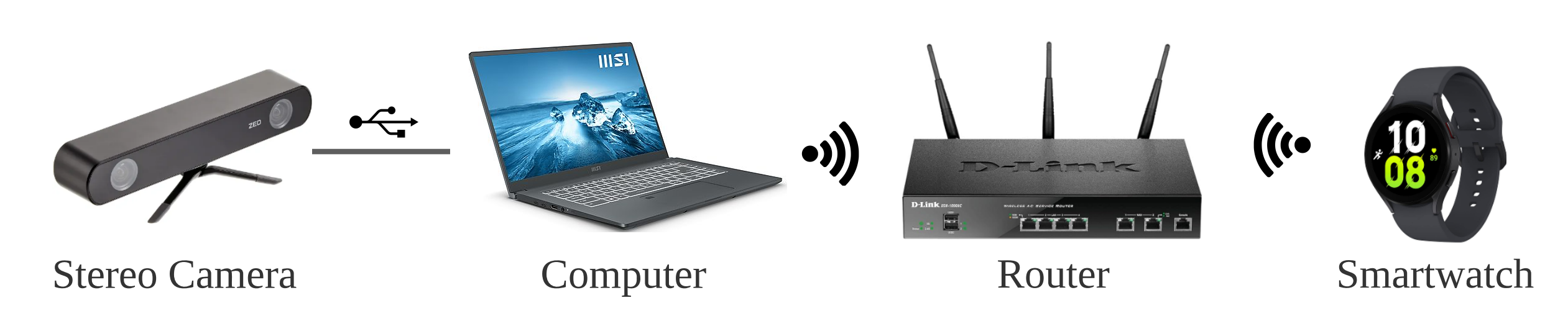}
    \caption{Diagram of the capture setup.}
    \label{fig:capsetup}
\end{figure}

In order to capture the data, a custom capturing rig has been developed. The diagram of such device is shown in Figure \ref{fig:capsetup}. Hardware-wise, it comprises four main components: an a computer, a stereo camera, a smartwatch and a router. The camera is connected through USB to the computer. Both computer and smartwatch are connected to the router via WiFi, which is the channel they use to share data.

Specifically, the components of the setup include an i7 processor with 16GB of RAM and a GTX 1650 GPU. This system is capable of running algorithms written in various programming languages and is equipped with WiFi, Bluetooth, wired Ethernet, USB 3.0 ports, and HDMI connections; a ZED, which is a stereo color camera capable of providing two images of 1080p resolution with a theoretical working rate of 30 Hz; and a Samsung Galaxy Watch 5, which features an accelerometer, gyroscope, and rotation vector, all running at 100Hz. The router is a D-Link DSR-1000AC with WiFi capabilities of up to 1300 Mbps.

\begin{figure}[H]
    \centering
    \includegraphics[width=0.8\textwidth]{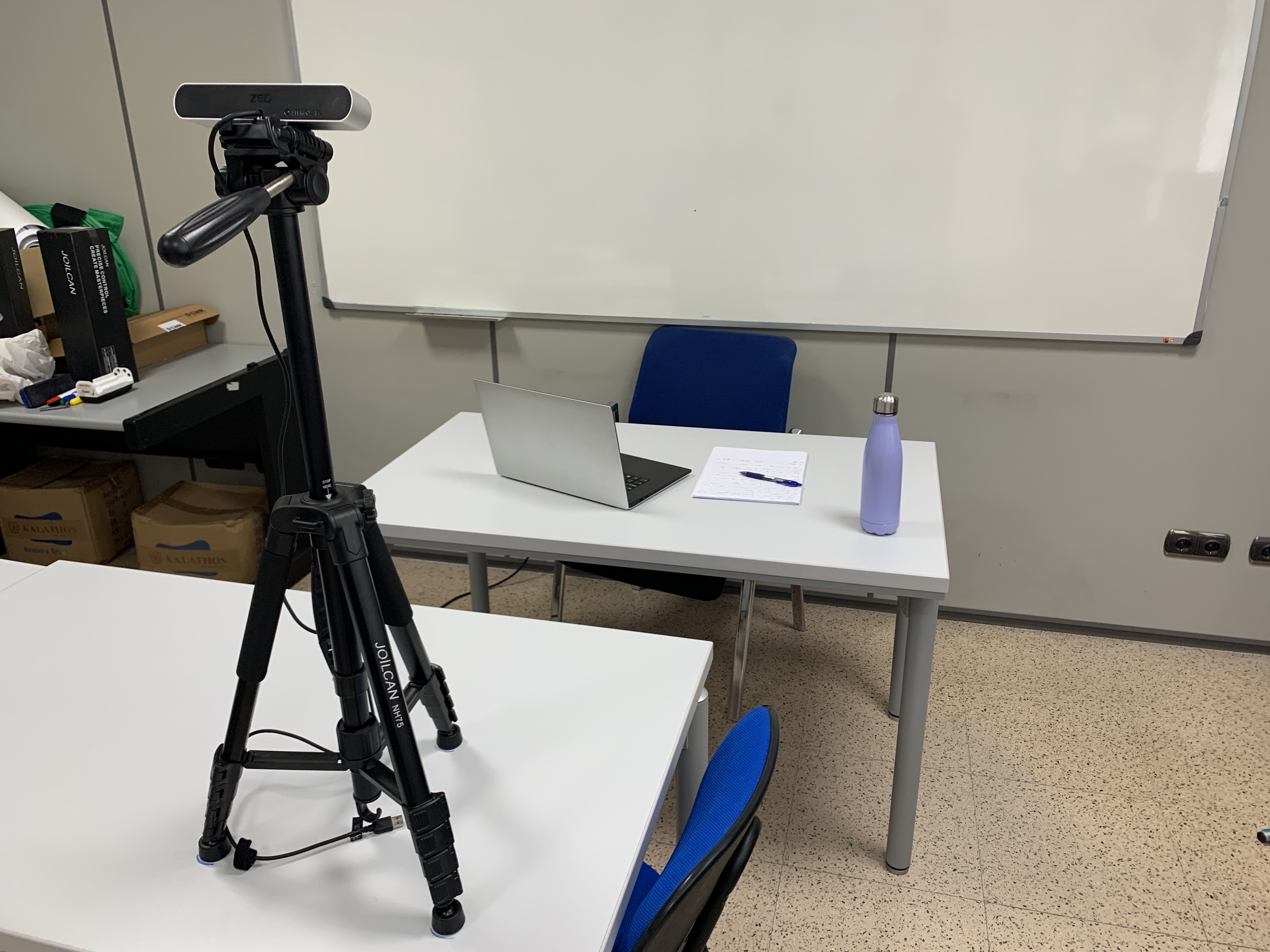}
    \caption{Capture setup.}
    \label{fig:setup}
\end{figure}

The setup, which is shown in Figure \ref{fig:setup}, also features a regular desk. The camera is placed on a tripod, in front of the user and pointed to them, a laptop, a wireless computer mouse, some pens and pencils, and a notebook or paper sheets. The users also carry their smartphone in their pocket, and a backpack with a bottle by their side placed on the floor. This setup aims to imitate the conditions of an actual student in a classroom. Due to this, the mentioned elements are organized by them with no further restrictions. The only requirement is that the camera is pointed towards them.

Regarding the software, the computer runs a custom program that accesses the stereo camera, directly through its USB connection, and stores the images. In parallel, the program also receives the data packets sent by the smartwatch, that runs a custom android app that gets the \Gls{imu} data and send them through the WiFi local area connection once per second.

\subsection*{Synchronization}
\label{subsec:synchro}

The dataset includes accelerometer data, gyroscope data, orientation data, and images. The accelerometer, gyroscope, and rotation vector records are provided by the smartwatch at a rate of 100MHz and are internally synchronized by the smartwatch itself. Thus, 100 records are obtained at the same time and packed with the same timestamp. The time of the computer is shared with the smartwatch so both are synchronized. The smartwatch time is triflingly delayed by the time it takes to the sync packet to travel from the computer to the smartwatch. The stereo camera is triggered by the computer that queries the images at 25-30 fps. Each pack of two images (left and right of the stereo setup) is labeled with the timestamp of the computer.

The smartwatch does not stream the \Gls{imu} data in real-time. Instead of that, the records are buffered in blocks of 100 readings in the smartwatch, so only one connection per second is done between the computer and the smartwatch. We do this to alleviate the computing requirements of the system and to avoid flooding the network. Images and \Gls{imu} data can be easily matched together as both computer and smartwatch are synchronized, and the records labeled with the corresponding timestamp. As expected, not all \Gls{imu} registers will have a corresponding image due to the difference in the working rates of the sensors.

\subsection*{Procedure}
\label{subsec:procedure}

The protocol we followed to record the data is as follows. First, the participants are instructed to sit down and prepare the desk with their laptop, notebook, mouse and pen and pencils. Their smartphone would be on their pocket and the backpack on the floor with the bottle in it. Each individual sets the environment on their own, as if they were in a real classroom.
Then, the procedure is different depending on the type of activity. For the continuous activity, the user is instructed to do it for 210 seconds. Only the central \textbf{200} seconds are actually stored, so we discard the first and last part of the sequence. Otherwise, if the activity to record is of type instantaneous, the participant is instructed to begin by the technician recording the data. This way, only the data related to the activity is recorded. In this case, each action is repeated 20 times, and takes the required time to perform it. This time may be different for each activity and each instance even within the same category.

As mentioned earlier, there are two types of activities: the continuous activities are those that are performed during an extensive timeframe. For instance, typing on smartphone, writing or drawing on a paper are within this category. On the other hand, the instantaneous activities are those that happen in a very narrowed time span, barely seconds. Raising a hand, drinking or stretching out are within this category. The list of activities considered in the dataset is shown in Table \ref{tab:activities}. As mentioned earlier, this dataset is intended to collect activities that are carried within a classroom during a regular lecture.

\section*{Data Records}
\label{sec:data-records}

%The Data Records section should be used to explain each data record associated with this work, including the repository where this information is stored, and to provide an overview of the data files and their formats. Each external data record should be cited numerically in the text of this section, for example \cite{Hao:gidmaps:2014}, and included in the main reference list as described below. A data citation should also be placed in the subsection of the Methods containing the data-collection or analytical procedure(s) used to derive the corresponding record. Providing a direct link to the dataset may also be helpful to readers (\hyperlink{https://doi.org/10.6084/m9.figshare.853801}{https://doi.org/10.6084/m9.figshare.853801}).

%Tables should be used to support the data records, and should clearly indicate the samples and subjects (study inputs), their provenance, and the experimental manipulations performed on each (please see 'Tables' below). They should also specify the data output resulting from each data-collection or analytical step, should these form part of the archived record.

In this section, we summarize the dataset, which includes various sensor data and stereo camera images captured during activities performed by participants. The data is organized in JSON format for sensors and PNG for images, with timestamps ensuring synchronization across all files. The dataset structure is designed to support the analysis of both continuous and instantaneous classroom activities.

\subsection*{Dataset Components}
\label{sec:dataset_components}

The dataset will store stereo camera images, labeled as A and B, along with sensor values captured by the smartwatch in JSON files.

As indicated in Table \ref{tab:sensor_data}, the sensor data reveals details about various sensors integrated into the system. The \texttt{Samsung HR Wakeup Sensor} measures the user's heart rate at a frequency of one value per second. The \texttt{Samsung Linear Acceleration} Sensor measures the device's linear acceleration in three axes, excluding gravity, with a sampling rate of 100 samples per second. The \texttt{LSM6DSO Gyroscope} reports the rate of rotation around its axes at the same sampling rate. \texttt{The Samsung Rotation Vector} sensor, with the same sampling rate, provides rotational information in three dimensions. The \texttt{OPT3007 Light Sensor} measures environmental light levels, expressing luminosity in lux units. 

%***frec cardica

The provided images have a resolution of 1920 x 1080 pixels, with pairs of images corresponding to left and right cameras in the stereo setup, each labeled with a timestamp. Parameters such as focus, exposure, gain, and white balance are automatically adjusted using built-in Image Signal Processor features. Images are stored in individual PNG files without compression. It is important to note that the dataset's images are neither rectified nor distorted in any way.

In total, the dataset comprises 472,03 minutes of recording, during which the following total number of samples has been collected:

\begin{itemize}
    \item Camera A Images: 437.904 images
    \item Camera B Images: 437.904 images
    \item Samsung HR None Wakeup Sensor: 31.727 samples
    \item Samsung Linear Acceleration Sensor: 3.172.700 samples
    \item LSM6DSO Gyroscope: 3.172.700 samples
    \item Samsung Rotation Vector: 3.172.700 samples
    \item OPT3007 Light: 208.157 samples
\end{itemize}

The frequency of these values collected by the watch in each experiment corresponds to the smartwatch's maximum data collection capacity. The frequency of the data corresponds to that shown in Table \ref{tab:sensor_data}.

The dataset was recorded using 12 participants. The Table \ref{tab:sota} summarizes the demographic and physical characteristics of the participants, including age, gender, height, weight, and hand laterality. These characteristics may influence the variability in the recorded activities and should be considered when analyzing the data.

\begin{figure}[H]
    \centering
    % Column titles
    \begin{minipage}[b]{0.45\textwidth}
        \centering
        \textbf{Continuous Gestures}
    \end{minipage}
    \hfill
    \begin{minipage}[b]{0.45\textwidth}
        \centering
        \textbf{Instantaneous Gestures}
    \end{minipage}
    \par\bigskip
    
    % Row 1
    \begin{subfigure}[b]{0.45\textwidth}
        \centering
        \includegraphics[width=\textwidth]{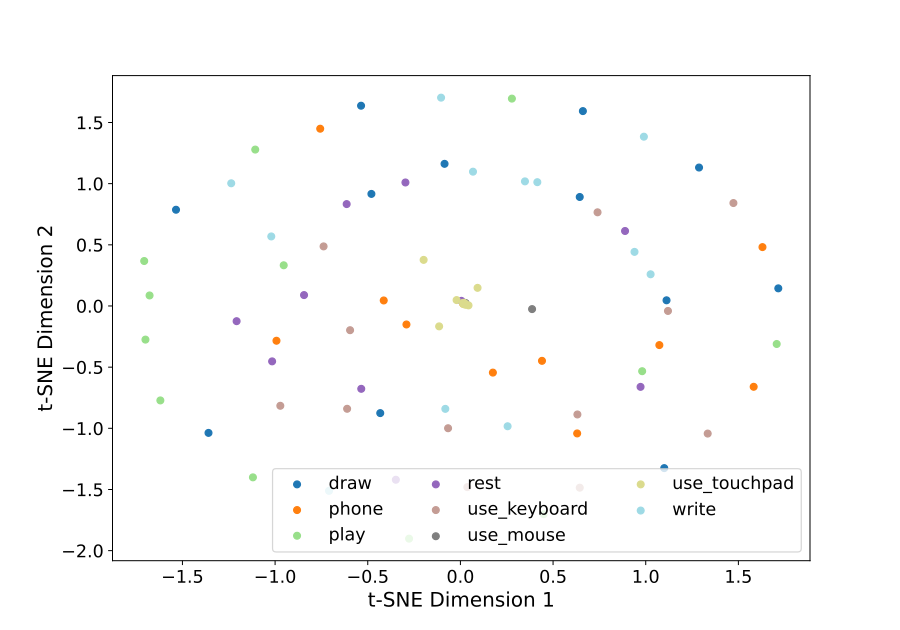}
        \caption{LSM6DSO Gyroscope Sensor Continuous}
        \label{fig:plot_inst_hr1}
    \end{subfigure}
    \hfill
    \begin{subfigure}[b]{0.45\textwidth}
        \centering
        \includegraphics[width=\textwidth]{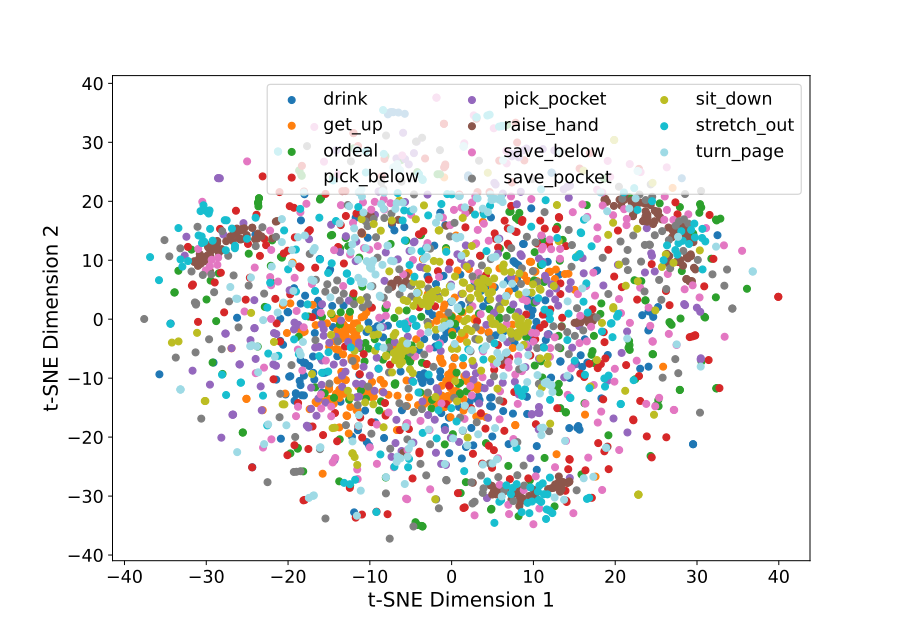}
        \caption{LSM6DSO Gyroscope Sensor Instantaneous}
        \label{fig:plot_inst_hr2}
    \end{subfigure}
    \par\bigskip
    
    % Row 2
    \begin{subfigure}[b]{0.45\textwidth}
        \centering
        \includegraphics[width=\textwidth]{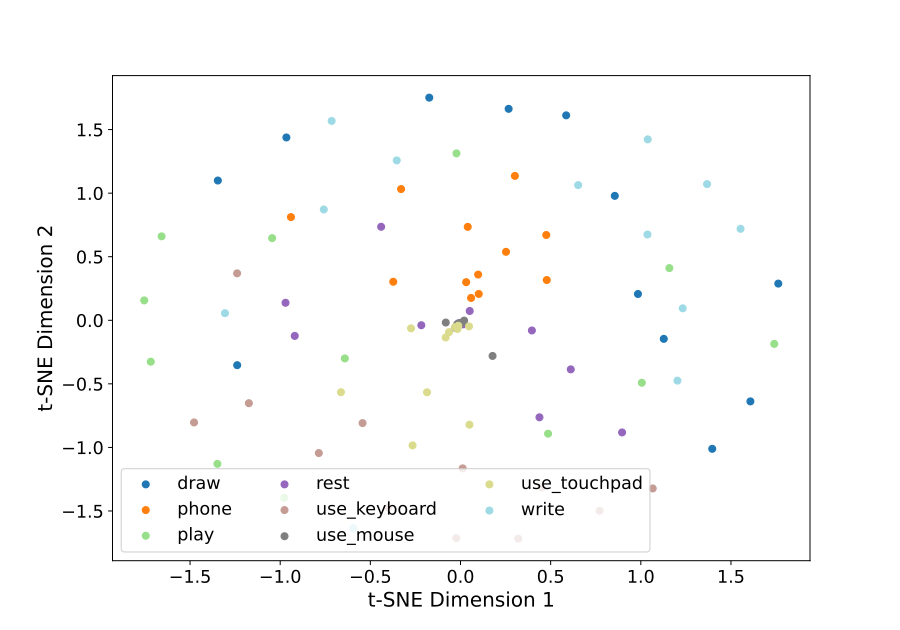}
        \caption{Samsung Linear Accelertion Sensor Continuous}
        \label{fig:plot_inst_hr3}
    \end{subfigure}
    \hfill
    \begin{subfigure}[b]{0.45\textwidth}
        \centering
        \includegraphics[width=\textwidth]{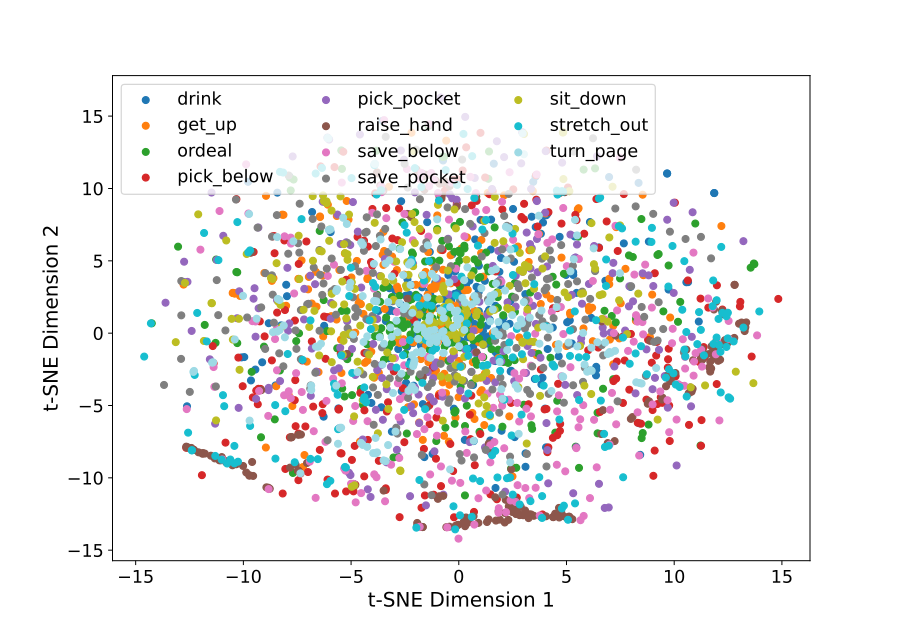}
        \caption{Samsung Linear Accelertion Sensor Instantaneous}
        \label{fig:plot_inst_hr4}
    \end{subfigure}
    \par\bigskip
    
    % Row 3
    \begin{subfigure}[b]{0.45\textwidth}
        \centering
        \includegraphics[width=\textwidth]{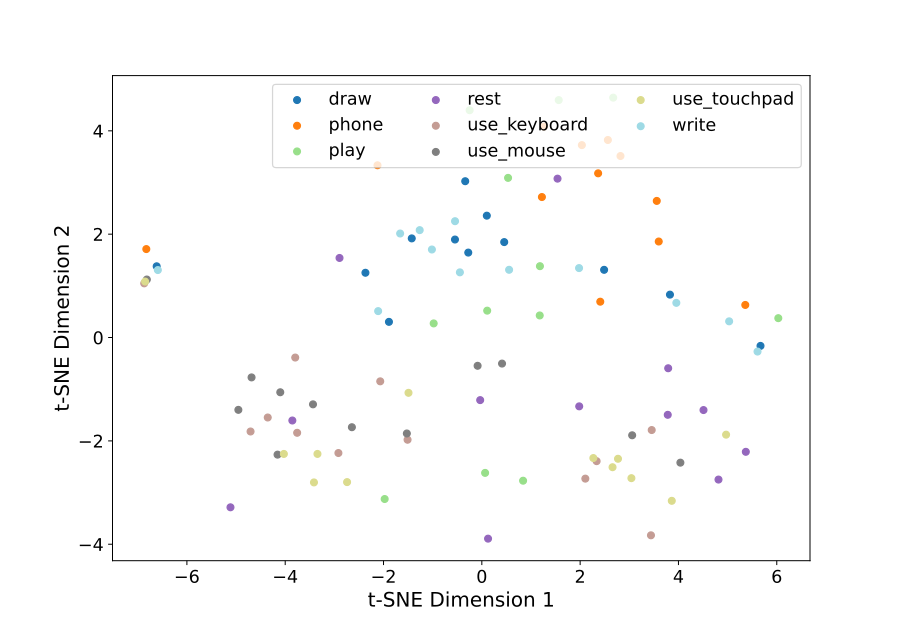}
        \caption{Rotation Vector Sensor Continuous}
        \label{fig:plot_inst_hr5}
    \end{subfigure}
    \hfill
    \begin{subfigure}[b]{0.45\textwidth}
        \centering
        \includegraphics[width=\textwidth]{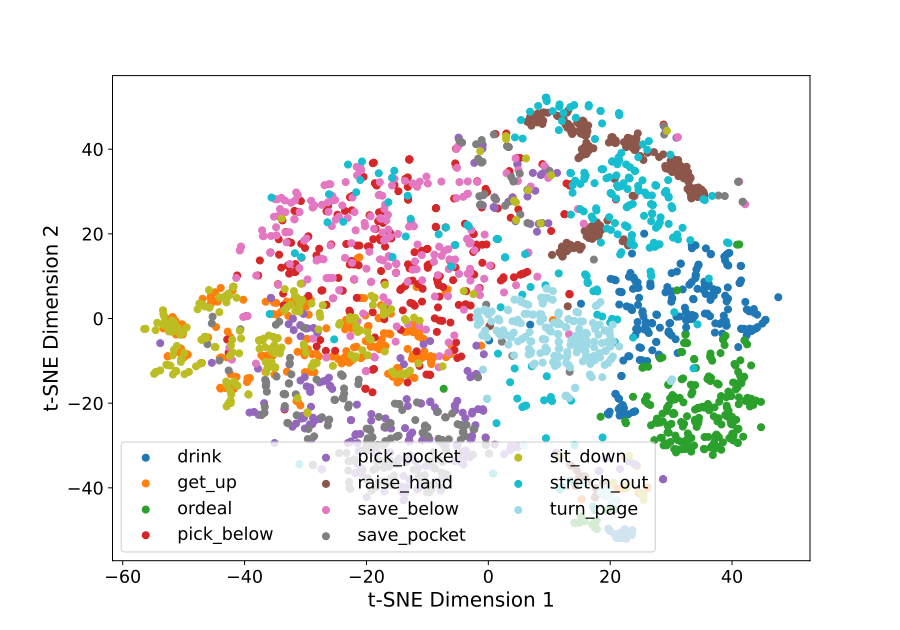}
        \caption{Rotation Vector Sensor Instantaneous}
        \label{fig:plot_inst_hr6}
    \end{subfigure}
    \par\bigskip
    
    \caption{Dimensionality reduction TSNE}
    \label{fig:plot_tsne}
\end{figure}

Additionally, Figure \ref{fig:plot_tsne} presents graphs illustrating non-linear dimensionality reduction using t-SNE. This technique enables the visualization of activities performed by each subject, highlighting clustering based on the sensor employed for analysis (Samsung Linear Acceleration, LSM6DSO Gyroscope, or Samsung Rotation Vector).

For discrete activities, each gesture by each subject is represented as a point in a 2D space, with missing values replaced by the mean. Conversely, for continuous activities, each class for each individual is depicted by a single point in the 2D space.

In these graphs, one can observe how certain variables cluster together and how some classes exhibit similarities. For instance, activities such as retrieving an item from a backpack or pocket, or between drawing and writing, show clear clustering.

\begin{figure}[H]
    \centering
    \begin{subfigure}[b]{0.8\textwidth}
        \centering
        \includegraphics[width=\textwidth]{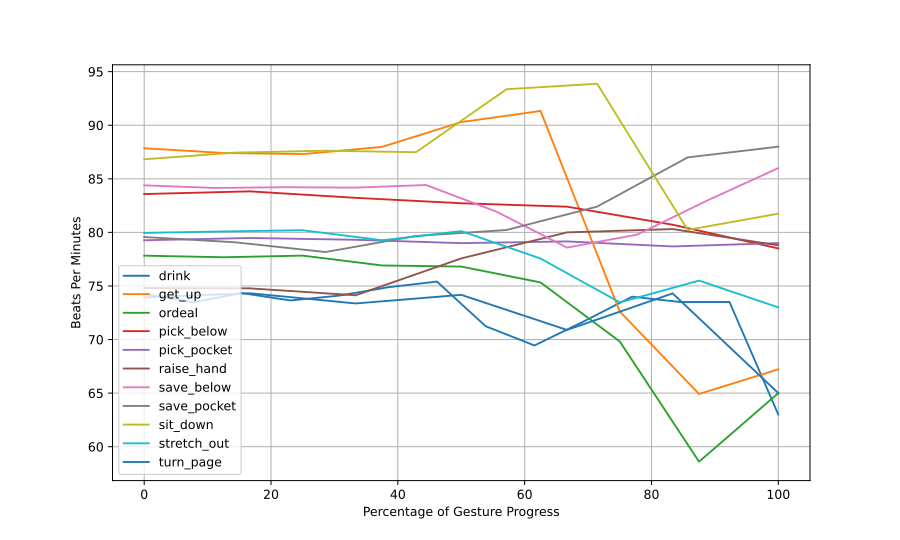}
        \caption{Evolution of heart rate during instantaneous gestures}
        \label{fig:plot_inst_hr_2}
    \end{subfigure}
    \par\bigskip
    \begin{subfigure}[b]{0.6\textwidth}
        \centering
        \includegraphics[width=\textwidth]{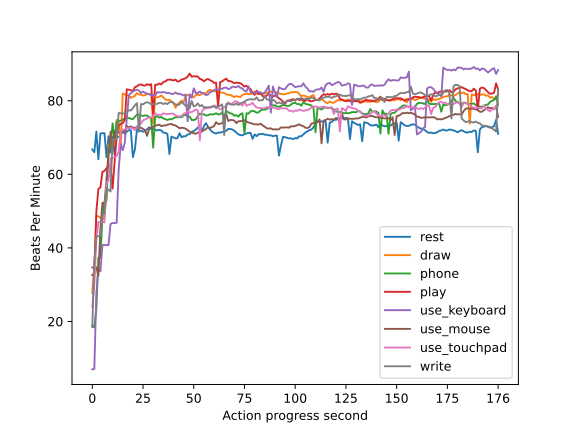}
        \caption{Evolution of heart rate during continuous gestures}
        \label{fig:plot_inst_hr_2_1}
    \end{subfigure}
    \caption{Evolution of heart rate by activity}
    \label{fig:heart_rate}
\end{figure}

Moreover, Figure \ref{fig:heart_rate} illustrates the evolution of heart rate. For continuous activities, the average heart rate of each subject for each activity has been calculated. For instantaneous activities, due to differences in duration, the data is presented as a percentage of the activity completed.

\subsection*{Organization}

The data collected from the SmartWatch is stored in JSON-formatted files, with one JSON file saved for each second of data collection for each user. These JSON files are stored with a title that reflects the hour, minute, second, and millisecond of a second of the data capture on the clock. To obtain the timestamp of each sensor's data capture moment, you need to examine the \texttt{timestamp} key of each record.

The captured images also include a timestamp of the moment of capture, facilitating their association with JSON files by searching within the subfolders of the experiment name.

The JSON structure consists of a key named \texttt{num\_samples}  indicating the number of samples of each sensor, along with a key named \texttt{data} containing the values of the watch sensors: \texttt{Samsung HR None Wakeup Sensor}, \texttt{Samsung Rotation Vector}, \texttt{LSM6DSO Gyroscope}, \texttt{Samsung Linear Acceleration Sensor}, and \texttt{OPT3007 Light}. Each of these sub-keys is a list of dictionaries with the precise time of the sensor measurement and the remaining values taken in that measurement.

\begin{figure}[H]
    \centering
    \includegraphics[width=0.4\textwidth]{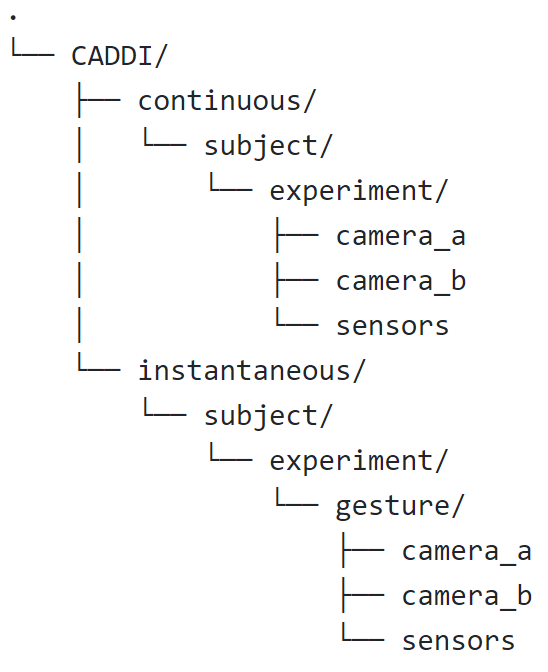}
    \caption{Directory Structure of the Dataset.}
    \label{fig:structure}
\end{figure}

Timestamps are displayed in the \texttt{HH:MM:SS.ff} format, indicating the time of day when the measurement was taken (UTC+1). The dataset directories are organized hierarchically. Initially, folders are categorized by experiment type, continuous or instantaneous. In both cases the following levels are subject, followed by experiment. In continuous, within each experiment there are the folders: camera\_a, camera\_b and sensors. In the instantaneous experiments, an additional level indicates the gesture number, with each gesture folder containing camera\_a, camera\_b, and sensors subfolders. The directory structure can be visualized in Figure \ref{fig:structure}.

\begin{enumerate}
    \item \texttt{CameraA}: Directory containing the image files of the left part of the stereo camera.
    \newline Path: \texttt{/type/subject/experiment/(gesture)/camera\_a/}
    \item \texttt{CameraB}: Directory containing the image files of the right part of the stereo camera.
    \newline Path: \texttt{/type/subject/experiment/(gesture)/camera\_b/}
    \item \texttt{Sensors}: Directory containing the sensors data in JSON files. 
    \newline Path: \texttt{/type/subject/experiment/(gesture)/sensors/}
\end{enumerate}

The data files (both dictionaries and PNG images) follow a naming convention based on the time of day when the image was captured or the biometric data was saved in format \texttt{HH\_MM\_SS\_ff}. The experiment names correspond to those shown in Table \ref{tab:activities}, adapted to snake case writing.

\section*{Technical Validation}
\label{sec:tech-val}
%This section presents any experiments or analyses that are needed to support the technical quality of the dataset. This section may be supported by figures and tables, as needed. This is a required section; authors must present information justifying the reliability of their data.

This dataset was validated using a deep learning model to classify activities. Data were segmented and split into training and validation sets to simulate performance on new users. The model showed higher accuracy for instantaneous actions, providing a useful baseline for future work with this dataset.

\subsection*{Benchmark}
\label{sec:benchmark}

To validate the IMU data, a deep learning model was trained to classify the 19 classes in the dataset. The input for the model was created by splitting the complete sequences into fixed-size segments using a sliding window approach \cite{imu}. The dataset was divided into training and validation sets by assigning a portion of the participants to each, ensuring that the model never sees any validation data during training. This also makes the results comparable to how the model would perform on entirely new users.

Out of the 12 participants in the dataset, 9 were used for training and 2 for validation. The data was segmented into 2-second windows with a 1-second overlap. The model architecture includes a 1D convolutional layer to extract local features and reduce dimensionality, followed by two LSTM layers to analyze the sequential data, and a dense layer for classification.

\begin{figure}[H]
    \centering
    \includegraphics[width=0.9\textwidth]{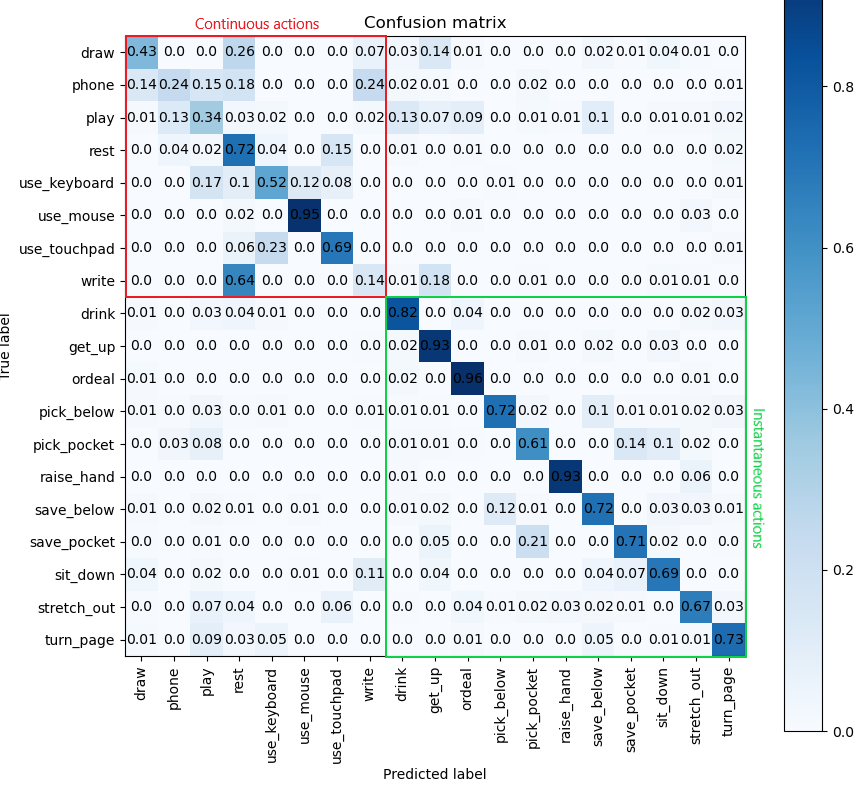}
    \caption{Confusion Matrix.}
    \label{fig:confusion_matrix}
\end{figure}

Using this model, a validation accuracy of around 65\% was achieved. Figure \ref{fig:confusion_matrix} shows the confusion matrix, which highlights that the model performs better on instantaneous actions, achieving an average accuracy of 77\%, compared to continuous actions, where the average accuracy is 45\%.

\section*{Usage Notes}

%The Usage Notes should contain brief instructions to assist other researchers with reuse of the data. This may include discussion of software packages that are suitable for analysing the assay data files, suggested downstream processing steps (e.g. normalization, etc.), or tips for integrating or comparing the data records with other datasets. Authors are encouraged to provide code, programs or data-processing workflows if they may help others understand or use the data. Please see our code availability policy for advice on supplying custom code alongside Data Descriptor manuscripts.

%For studies involving privacy or safety controls on public access to the data, this section should describe in detail these controls, including how authors can apply to access the data, what criteria will be used to determine who may access the data, and any limitations on data use. 

This dataset is designed to facilitate the analysis of in-class activities using IMU data from low-cost sensors. Researchers can utilize this dataset for developing and testing algorithms that detect and predict student activities in a classroom setting. The following notes provide guidelines to aid in the effective reuse of the data:

\begin{itemize}
    \item \textbf{Software and Tools}:
    \begin{itemize}
        \item Recommended software packages for analyzing the IMU data include Python libraries such as \texttt{numpy}, \texttt{pandas}, and \texttt{scikit-learn} for data manipulation and machine learning, as well as \texttt{tensorflow} or \texttt{pytorch} for deep learning models.
        \item Visualization of the data can be performed using \texttt{matplotlib} and \texttt{seaborn}.
    \end{itemize}

    \item \textbf{Integration with Other Datasets}:
    \begin{itemize}
        \item To compare or integrate this dataset with other similar datasets, ensure that the data formats and labeling conventions are aligned.
        \item Researchers can use common benchmarks and metrics for activity recognition to evaluate the performance of their models across different datasets.
    \end{itemize}

    \item \textbf{Privacy and Data Access}:
    \begin{itemize}
        \item This dataset does not contain personally identifiable information, except for the recorded faces, thus minimizing privacy concerns. However, any further distribution of the dataset must comply with relevant data protection regulations. All participants have given their consent for the publication of the dataset.
    \end{itemize}

    \item \textbf{Code Availability}:
    \begin{itemize}
        \item Example code for preprocessing is provided in code avliblity \ref{sec:code_availability}. Researchers are encouraged to build upon these examples to suit their specific needs.
        \item Additional data-processing workflows or scripts can be made available upon request.
    \end{itemize}
\end{itemize}

By following these guidelines, researchers can maximize the utility of the dataset and contribute to the advancement of activity recognition technologies in educational settings.

\section*{Code Availability}
\label{sec:code_availability}
%For all studies using custom code in the generation or processing of datasets, a statement must be included under the heading "Code availability", indicating whether and how the code can be accessed, including any restrictions to access. This section should also include information on the versions of any software used, if relevant, and any specific variables or parameters used to generate, test, or process the current dataset. 

In this section, we provide details on the accessibility and code used to process the data. The code can be accessed through the following link: \href{https://bitbucket.org/rovitlib/caddi-an-in-class-activity-detection-dataset-using-imu-data/src/main/}{CADDI: An In-class Activity Detection Dataset Using IMU Data}.

The provided code generates a DataFrame where each entry contains information about a complete sequence of an experiment, i.e., data recorded sequentially. For continuous data, each experiment is contained in a single entry, and for instantaneous data, each entry contains information from a repetition of an experiment. An entry in the dataset consists of the following parts:

\begin{itemize}
    \item \textbf{Type:} Type of experiment, continuous or instantaneous.
    \item \textbf{Subject:} The subject performing the action.
    \item \textbf{Action:} Action performed in the experiment.
    \item \textbf{Gesture:} Indicator of the experiment repetition number.
    \item \textbf{Path\_Camera\_A:} Path to the left camera data.
    \item \textbf{Path\_Camera\_B:} Path to the right camera data.
    \item \textbf{Path\_Sensors:} Path to the sensor files in JSON format.
    \item \textbf{Camera\_A:} List of names of the left camera photos.
    \item \textbf{Camera\_B:} List of names of the right camera photos.
    \item \textbf{Sensors:} List of names of the sensor files in JSON format.
\end{itemize}

Along with the code that generates the DataFrame, two functions are added to process the content of a JSON file.

\begin{figure}[H]
    \centering
    \includegraphics[width=0.5\textwidth]{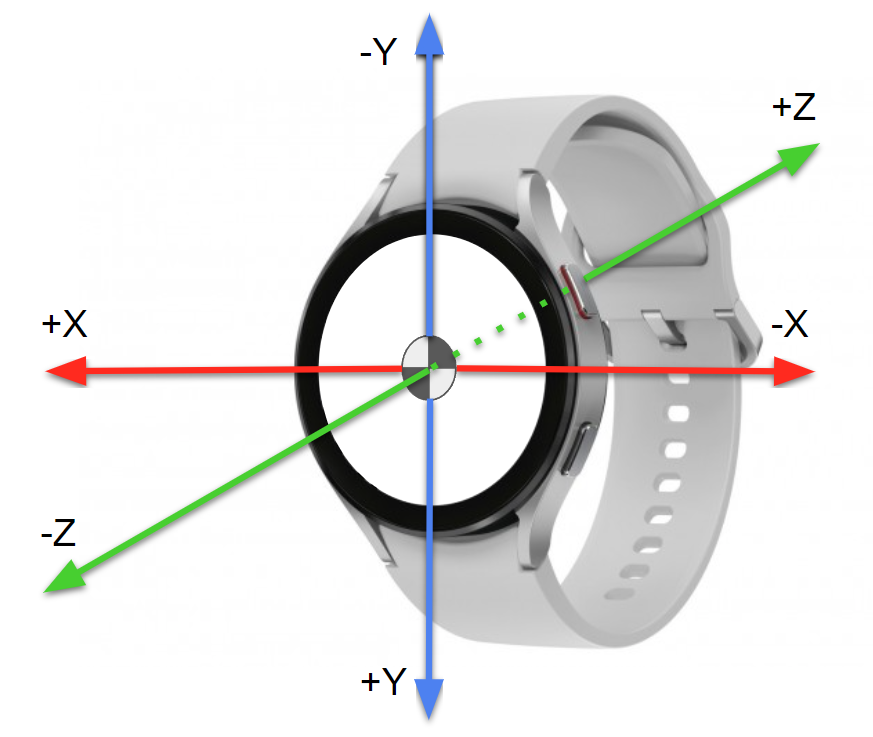}
    \caption{Smartwatch Axes in Sensor Readings.}
    \label{fig:axes}
\end{figure}

The first function, given the path of a JSON, reads and generates a DataFrame with the information from the file. This DataFrame contains 100 entries, one for each time inertial sensors capture information per second (100 Hz). Each entry contains precise timing information when the sensor data was captured and the values in X, Y, and Z coordinates for the accelerometer and gyroscope, and a quaternion X, Y, Z, and W for the rotation vector (\texttt{Acc\_Time}, \texttt{Acc\_X}, \texttt{Acc\_Y}, \texttt{Acc\_Z}, \texttt{Gyro\_Time}, \texttt{Gyro\_X}, \texttt{Gyro\_Y}, \texttt{Gyro\_Z}, \texttt{Rot\_Time}, \texttt{Rot\_X}, \texttt{Rot\_Y}, \texttt{Rot\_Z}, \texttt{Rot\_W}). In Figure \ref{fig:axes}, the orientation of each axis on the smartwatch can be observed.

The second function generates a \textit{numpy array} of shape (100, 10), where the first dimension corresponds to the values taken in one second and the second to the sensor values excluding the time stamps (100 values * \texttt{Acc\_X}, \texttt{Acc\_Y}, \texttt{Acc\_Z}, \texttt{Gyro\_X}, \texttt{Gyro\_Y}, \texttt{Gyro\_Z}, \texttt{Rot\_X}, \texttt{Rot\_Y}, \texttt{Rot\_Z}, \texttt{Rot\_W}). A version of this function analyzes all JSON files from a directory and generates an array of shape (\textit{X}, 100, 10) where \textit{X} is the number of files in the directory.

\begin{figure}[H]
    \centering
    \includegraphics[width=0.7\textwidth]{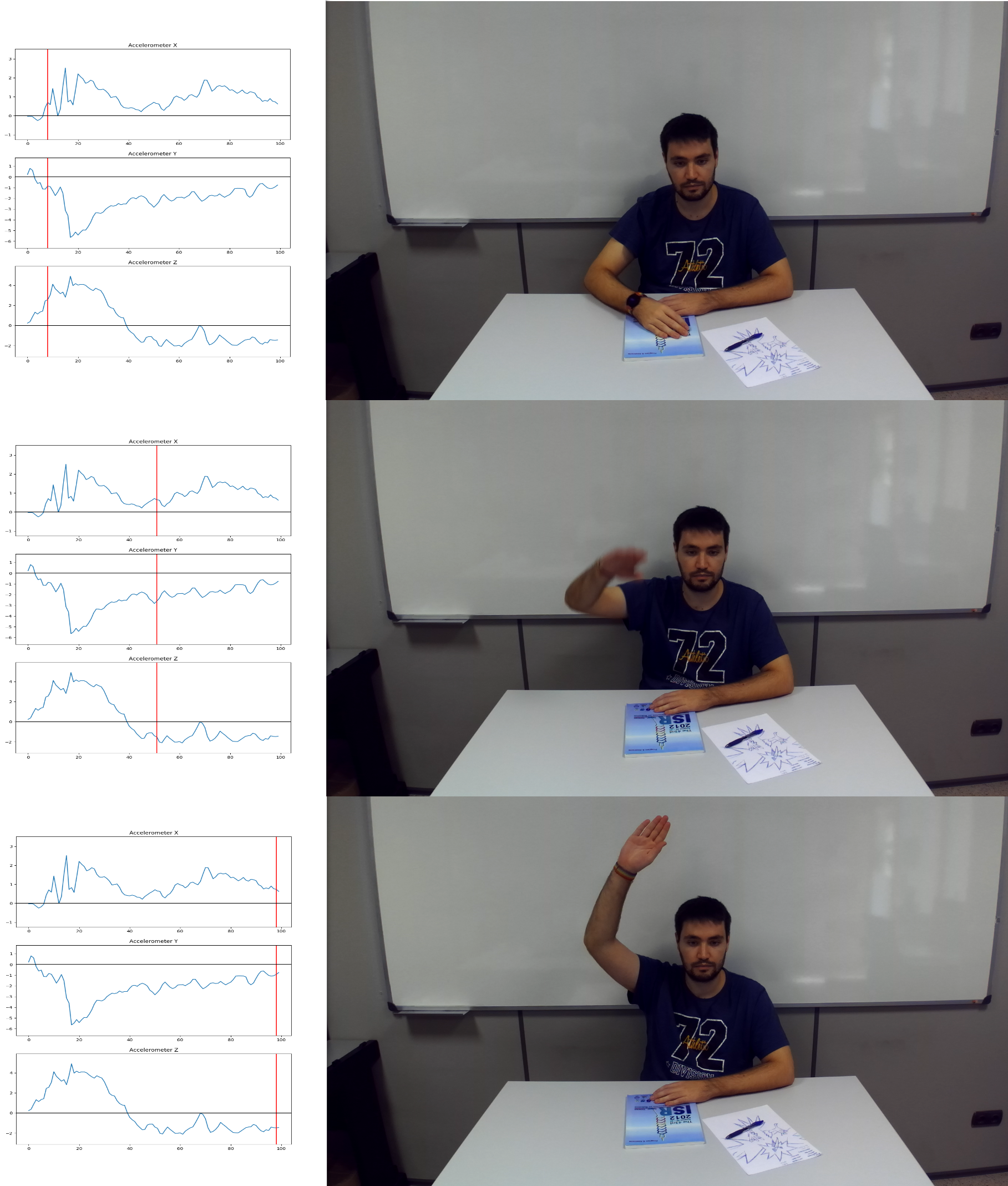} % Adjust width as needed
    \caption{Data Visualizer for Smartwatch Sensors and Image Data}
    \label{fig:visualizer}
\end{figure}

Additionally, a function is included that enables visualization of the contents of a DataFrame entry. This function displays the sequence of images from Camera A alongside a diagram with accelerometer data. The visualization helps in understanding the sequence and dynamics of the experiment based on the captured sensor and image data. An example of how the viewer works can be seen in Figure \ref{fig:visualizer}

%\noindent LaTeX formats citations and references automatically using the bibliography records in your .bib file, which you can edit via the project menu. Use the cite command for an inline citation, e.g. \cite{Kaufman2020, Figueredo:2009dg, Babichev2002, behringer2014manipulating}. For data citations of datasets uploaded to e.g. \emph{figshare}, please use the \verb|howpublished| option in the bib entry to specify the platform and the link, as in the \verb|Hao:gidmaps:2014| example in the sample bibliography file. For journal articles, DOIs should be included for works in press that do not yet have volume or page numbers. For other journal articles, DOIs should be included uniformly for all articles or not at all. We recommend that you encode all DOIs in your bibtex database as full URLs, e.g. https://doi.org/10.1007/s12110-009-9068-2.

\section*{Acknowledgements}

%Acknowledgements should be brief, and should not include thanks to anonymous referees and editors, or effusive comments. Grant or contribution numbers may be acknowledged.

Grant CIPROM/2021/17 funded by Prometeo program from Conselleria de Innovación, Universidades, Ciencia y Sociedad Digital of Generalitat Valenciana (Spain). The study was conducted in accordance with the Declaration of Helsinki. All participants signed an informed consent previous to the experiments. The dataset has been made publicly available under the CC-BY license.

\section*{Author contributions statement}

%Must include all authors, identified by initials, for example:
%A.A. conceived the experiment(s), A.A. and B.A. conducted the experiment(s), C.A. and D.A. %analysed the results. All authors reviewed the manuscript. 

L. M.: Software, Validation, Data Curation. S. S.: Software, Formal analysis, Data Curation. M. P.: Methodology, Visualization, Software. M. C.: Resources, Writing - Review \& Editing, Funding acquisition. F. G.: Conceptualization, Writing - Original Draft, Supervision.

\section*{Competing interests} 
%The corresponding author is responsible for providing a \href{https://www.nature.com/sdata/policies/editorial-and-publishing-policies#competing}{competing interests statement} on behalf of all authors of the paper. This statement must be included in the submitted article file.

The authors declare no known competing financial interests.

\section*{Figures \& Tables}

% Diagrama del set de grabación

% Estructura de directorios

% TSNE visualizations

% Ensures that all figures are placed before the tables
\clearpage 

% Tablas

\begin{table}[H]
\centering
\begin{tabular}{|p{5cm}|p{5cm}|}
\hline
 \multicolumn{1}{|c|}{Continuous Activity} & \multicolumn{1}{c|}{Instantaneous Activity} \\ \hline
Typing on smartphone    &    Raise a hand                      \\ \hline
Writing on a paper      &    Drink                             \\ \hline
Drawing on a paper      &    Stretch out                       \\ \hline
Typing on a keyboard    &    Pick from below                   \\ \hline
Using a mouse           &    Pick from pocket                  \\ \hline
Using a touchpad        &    Get up                            \\ \hline
Rest                    &    Sit down                          \\ \hline
Playing with a pen      &    Turn a page                       \\ \hline
                        &    Save below                        \\ \hline
                        &    Save pocket                       \\ \hline
                        &    Ordeal                            \\ \hline
\end{tabular}
\caption{\label{tab:activities}Two sets of continuous and instantaneous activities were recorded by all the participants. The first set are actions that happen during a big time frame whilst the second comprises events that span a small time frame.}
\end{table}

\begin{table}[H]
\centering
\resizebox{\textwidth}{!}{%
\begin{tabular}{|p{3cm}|p{5cm}|p{5cm}|p{2cm}|p{3.5cm}|}
\hline
\textbf{Official Sensor Name} & \textbf{Description} & \textbf{Measured Values} & \textbf{Samples per Second} & \textbf{Units in Data Sampling} \\
\hline
Samsung HR None Wakeup Sensor & Detects the moment the smartwatch awakens & beats per minute & 1 & BPM \\
\hline
Samsung Linear Acceleration Sensor & Reports linear acceleration excluding gravity in the sensor frame. & 
Value 0: X-axis, Value 1: Y-axis, Value 2: Z-axis & 100 & 
Value 0: m/s\textsuperscript{2}, Value 1: m/s\textsuperscript{2}, Value 2: m/s\textsuperscript{2} \\
\hline
LSM6DSO Gyroscope & Reports the rate of rotation around three sensor axes. & 
Value 0: rad/s, Value 1: rad/s, Value 2: rad/s & 100 & 
Value 0: m/s\textsuperscript{2}, Value 1: m/s\textsuperscript{2}, Value 2: m/s\textsuperscript{2} \\
\hline
OPT3007 Light Sensor & Measures the level of environmental light. & Luminosity value & 5 & Lux units \\
\hline
Samsung Rotation Vector & Provides continuous information about device orientation in three-dimensional space. & 
Value 0: X-axis vector $\ast$ sin($\theta$/2), Value 1: Y-axis vector $\ast$ sin($\theta$/2), Value 2: Z-axis vector $\ast$ sin($\theta$/2), Value 3: cos($\theta$/2)
 & 100 & No units (quaternion) \\
\hline
\end{tabular}}
\caption{\label{tab:sensor_data}Description of various sensors with their respective data values.}
\end{table}

\begin{table}[H]
\centering
\resizebox{1\textwidth}{!}{%
\begin{tabular}{|l|c|>{\centering\arraybackslash}p{2cm}|c|c|c|c|>{\centering\arraybackslash}p{2cm}|}
\hline
\multicolumn{1}{|c|}{Subject ID} & Hand Laterality & Watch Placement on Hand & Age & Gender & Height (cm) & Weight (Kg) \\ \hline
Subject 1    & Right & Right       & 24  & Male      & 169cm & 66kg   \\ \hline
Subject 2    & Right & Right       & 26  & Male      & 175cm & 65kg   \\ \hline
Subject 3    & Right & Right/left  & 23  & Male      & 185cm & 90kg   \\ \hline
Subject 4    & Right & Right       & 26 & Female     & 170cm & 55kg   \\ \hline
Subject 5    & Right & Right       & 22 & Male       & 172cm & 69kg   \\ \hline
Subject 6    & Right & Right       & 33 & Male       & 175cm & 95kg   \\ \hline
Subject 7    & Left  & Left        & 21 & Male       & 180cm & 82kg   \\ \hline
Subject 8    & Right & Right       & 38  & Male      & 167cm & 73kg   \\ \hline
Subject 9    & Right & Right       & 27  & Male      & 186cm & 75kg   \\ \hline
Subject 10   & Right & Right       & 24  & Female    & 170cm & 72kg   \\ \hline
Subject 11   & Right & Right       & 23  & Male      & 185cm & 100kg  \\ \hline
Subject 12   & Right & Right       & 25  & Male      & 170cm & 70kg   \\ \hline
\end{tabular}}
\caption{\label{tab:sota}Physical characteristics of the subjects in the dataset.}
\end{table}


\begin{thebibliography}{99}

\bibitem{ada_alevizaki_2022_7092553}
Ada Alevizaki and Niki Trigoni, 
\textit{watchHAR: A Smartwatch IMU dataset for Activities of Daily Living}, 
Zenodo, Sep 2022. 
DOI: \href{https://doi.org/10.5281/zenodo.7092553}{10.5281/zenodo.7092553}.

\bibitem{MartnezZarzuela2023VIDIMUMV}
Mario Martínez-Zarzuela, Javier González-Alonso, Míriam Antón-Rodríguez, Francisco Javier Díaz Pernas, Henning Müller, and Cristina Simon-Martinez, 
\textit{VIDIMU. Multimodal video and IMU kinematic dataset on daily life activities using affordable devices}, 
ArXiv, 2023. 
URL: \href{https://api.semanticscholar.org/CorpusID:257771819}{Semantic Scholar}.

\bibitem{misc_pamap2_physical_activity_monitoring_231}
Attila Reiss, 
\textit{PAMAP2 Physical Activity Monitoring}, 
UCI Machine Learning Repository, 2012. 
DOI: \href{https://doi.org/10.24432/C5NW2H}{10.24432/C5NW2H}.

\bibitem{chan2021a}
S. Chan Chang, R. Walmsley, J. Gershuny, T. Harms, E. Thomas, K. Milton, P. Kelly, C. Foster, A. Wong, N. Gray, S. Haque, S. Hollowell, and A. Doherty, 
\textit{Capture-24: Activity tracker dataset for human activity recognition}, 
University of Oxford, 2021.

\bibitem{misc_mhealth_dataset_319}
Oresti Banos, Rafael Garcia, and Alejandro Saez, 
\textit{MHEALTH}, 
UCI Machine Learning Repository, 2014. 
DOI: \href{https://doi.org/10.24432/C5TW22}{10.24432/C5TW22}.

\bibitem{misc_opportunity_activity_recognition_226}
Daniel Roggen, Alberto Calatroni, Long-Van Nguyen-Dinh, Ricardo Chavarriaga, and Hesam Sagha, 
\textit{OPPORTUNITY Activity Recognition}, 
UCI Machine Learning Repository, 2010. 
DOI: \href{https://doi.org/10.24432/C5M027}{10.24432/C5M027}.

\bibitem{yax2-ge53-21}
Mathias Ciliberto, Vitor Fortes Rey, Alberto Calatroni, Paul Lukowicz, and Daniel Roggen, 
\textit{Opportunity++: A Multimodal Dataset for Video- and Wearable, Object and Ambient Sensors-based Human Activity Recognition}, 
IEEE Dataport, 2021. 
DOI: \href{https://dx.doi.org/10.21227/yax2-ge53}{10.21227/yax2-ge53}.

\bibitem{https://doi.org/10.4121/12716081.v2}
Stylianos Paraschiakos, Marian Beekman, Arno Knobbe, Ricardo Cachucho, and Eline Slagboom, 
\textit{GOTOV Human Physical Activity and Energy Expenditure Dataset on Older Individuals}, 
4TU.ResearchData, 2021. 
DOI: \href{https://doi.org/10.4121/12716081.V2}{10.4121/12716081.V2}.

\bibitem{misc_realdisp_activity_recognition_dataset_305}
Oresti Banos, Mate Toth, and Oliver Amft, 
\textit{REALDISP Activity Recognition Dataset}, 
UCI Machine Learning Repository, 2014. 
DOI: \href{https://doi.org/10.24432/C5GP6D}{10.24432/C5GP6D}.

\bibitem{misc_human_activity_recognition_using_smartphones_240}
Jorge Reyes-Ortiz, Davide Anguita, Alessandro Ghio, Luca Oneto, and Xavier Parra, 
\textit{Human Activity Recognition Using Smartphones}, 
UCI Machine Learning Repository, 2012. 
DOI: \href{https://doi.org/10.24432/C54S4K}{10.24432/C54S4K}.

\bibitem{imu}
Sakorn Mekruksavanich and Anuchit Jitpattanakul, 
\textit{Deep Convolutional Neural Network with RNNs for Complex Activity Recognition Using Wrist-Worn Wearable Sensor Data}, 
Electronics, vol. 10, 2021, p. 1685. 
DOI: \href{https://doi.org/10.3390/electronics10141685}{10.3390/electronics10141685}.

\end{thebibliography}
\end{document}